# Lateral Stability of Vehicle with Interlocking Spikes

Volker Nannen[1] and Damian Bover[2]

[1]Sedewa, Finca Ecologica Son Duri, Vilafranca de Bonany, Spain.
Email: vnannen@gmail.com
[2]Sedewa, Finca Ecologica Son Duri, Vilafranca de Bonany, Spain.
Email: utopusproject@gmail.com

**ABSTRACT**

The interlock drive system generates traction by penetrating narrow articulated spikes into the ground and by using the strength of the deeper soil layers to resist horizontal draft forces. The system promises good tractive performance in low gravity environments where tires generate little traction due to low vehicle weight. Possible applications include heavy-duty vehicles for civil engineering tasks like earthmoving or mining excavation. Safe vehicle operation in complex terrain geometry requires lateral vehicle stability to prevent vehicle rollover. Good lateral stability is a particular requirement for excavation and piling operations where the margins of safety define the terrain geometry that can be worked in, and it is a major constraint in operational planning. An earthmoving vehicle that can operate at a high roll angle reduces the need to maintain ramps in pits and on piles and can shorten and simplify the paths for individual maneuvers. Here we report on several field trials on the lateral stability of an earthmoving vehicle that uses the interlock drive systems. We find that the vehicle can work well at roll angles of up to 20°, but that it needs further improvement if work at roll angles of 30° or more is required.

**INTRODUCTION**

Renewed government and commercial interest in lunar and planetary mining and construction pose the challenging engineering question of how to excavate, transport, and tip granular materials—a field long dominated by heavy machinery—in the low gravity environments of the Moon, Mars, and the larger asteroids. Mueller and King (2008) identify the following basic requirements for lunar excavators: navigate the lunar surface without getting stuck, avoid rocks, traverse 20° slopes when fully loaded, minimize total power and peak power, operate reliably over long periods. Just et al. (2020) identify excavation rate, traverse speed, power consumption, and simulant properties as the key performance metrics for the evaluation of regolith excavation techniques. Excavation rate, traverse speed, and power consumption depend on the terrain and the geometry of the deposit. In terrain that is not flat and does not have dedicated access roads, a vehicle that can move or work at high pitch or roll angles can travel shorter distances, which increases traverse speed and reduces vehicle power consumption. Reducing the need to create and maintain a system of access roads also simplifies the logistics and reduces the overall power consumption of mining and other civil engineering projects. The ability of a machine to move and work on a slope angle of at least 20° is therefore a key parameter for the selection of suitable excavation techniques.

Tires generate traction on granular material through friction with and between the upper particles of the material on which they move. On terrestrial soils, tires achieve an optimal tractive efficiency near a pull/weight ratio of 0.4 (Zoz and Grisso, 2003). In a lunar environment, Wilkinson and DeGennaro (2007) expect a lower pull/weight ratio of 0.21. This poses a serious challenge to the design of civil engineering vehicles like rippers, scrapers, and crawlers. If they rely on tires for traction, every kg of launch mass is expected to generate a tractive force of only 0.35 N.

An alternative traction method is to interlock spikes with the ground (Bover 2011; Nannen et al. 2016, 2017). As described by Nannen and Bover (2021), efficient and reliable ground penetration can be achieved if narrow spikes are attached to a lever arm that is attached to a hinge close to the ground, see Figure 1. A backward force on such a spike drives it into the ground to a depth where the lateral strength of the ground equals the draft force, in a self-regulating manner. A forward force pulls the spike out of the ground. Such interlocking spikes can be integrated with a push-pull vehicle (Creager et al., 2012) where alternating frames push or pull tools like rippers or blades from the anchored spikes.

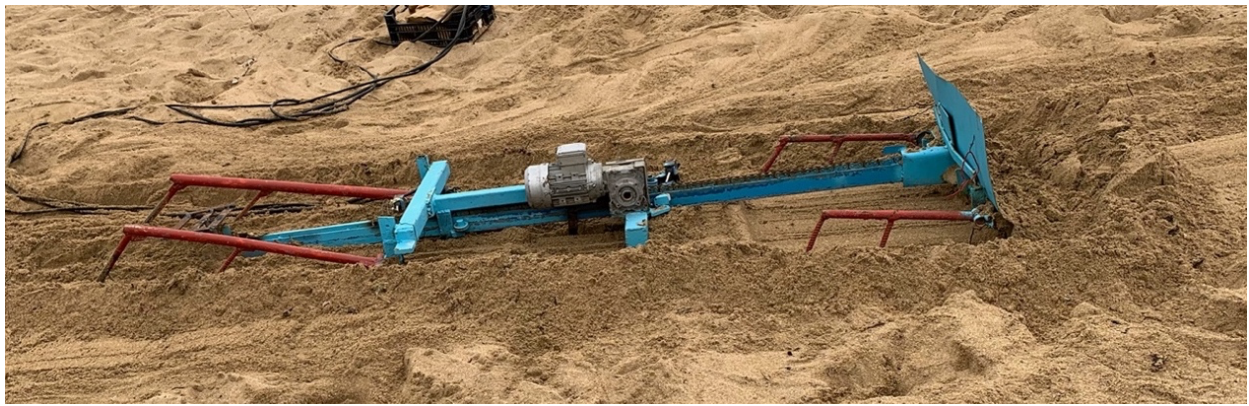

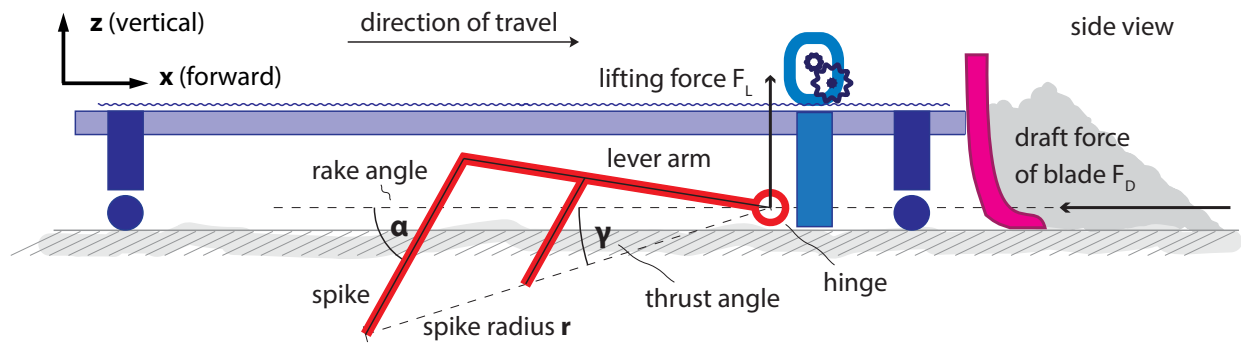

Figure 1: **Crawler with spikes for traction. Top: demonstrator on beach sand. The spikes are painted red. In an alternating motion pattern, the two spikes on the rear frame on the left push the blade forward, and the two spikes attached to the blade on the right pull the rest of the vehicle forward. Bottom: schematic drawing. Only one of the spikes that push the blade is shown.**

Important design parameters for the spike are the diameter of the spike, the rake angle $\alpha$ between the soil surface and the spike, and the thrust angle $\gamma$ between the soil surface and a line from the spike tip to the hinge. $\gamma$ in turn depends on the spike radius r, which is the distance between spike tip and hinge, as well as the elevation of the hinge above the ground. Spike diameter and rake angle determine the resistive force of the soil. The difference $\alpha - \gamma$ defines the penetration force and is best kept in the range $15° < \alpha - \gamma < 35°$. See Table 1 for a list of symbols.

**Table 1: List of symbols**

| | |
|---|---|
| α | rake angle, inclination from horizontal of the spike |
| γ | thrust angle, inclination from horizontal of the line from hinge to spike tip |
| $F_D$ | draft force at the blade |
| $F_L$ | lift force at the hinge |
| r | spike radius, the distance between spike tip and hinge |

Since a horizontal draft force $F_D$ creates a lift force $F_L = F_D \times \tan \gamma$ at the hinge, keeping the thrust angle small is important for vehicle stability. For a vehicle with an intended pull/weight ratio of 2, the thrust angle γ should not exceed 26.5°. With spikes as described here, if more tractive force is required from a vehicle with a given weight, the pull/weight ratio can be improved by increasing the spike radius. There is no need to add ballast.

To test the ability of a push-pull vehicle that uses interlocking spikes for traction to move and work on a slope of granular material, we performed field trials with a crawler in two relevant environments: a quarry and a beach. The task of the crawler was defined as path clearing, i.e., to clear away ridges and fill gaps in the ground, so that a smooth path is created.

**EXPERIMENTAL SETUP**

We conducted the field trials on fine-medium beach sand of bioclastic origin at the Bay of Palma, and on fine to coarse granular material at a stone quarry near the town of Petra, both on the island of Mallorca, Spain.

The crawler was small and light enough to be deployed manually. The blade was 80 cm wide and 40 cm high, forming the widest and highest part of the vehicle. The blade was welded to the crawler, with a fixed vertical position and angle. The total vehicle weight was 40 kg. The total vehicle length without spikes was 2.2 m. The center of mass of the crawler was about 10 cm above the ground. The crawler consisted of two alternating frames. The first frame consisted of a 2 m long central bar, with the blade at its front and two small spikes to the rear of the blade, 60 cm apart. The second frame consisted of a 60 cm bar orthogonal to and centered at the long bar, to which an electric motor with a gearbox and two large spikes were attached, 55 cm apart. The motor drove a cogwheel that meshed with a motorcycle chain welded to the top of the central bar. By moving the second frame back and forth along the central bar, the two pairs of spikes were pushed into and pulled out of the soil in an alternating motion pattern, pushing the crawler forward. The second frame traveled 1.15 m along the central bar. The crawler was powered via cable by solar cells, backed up by a car battery for stability. Peak power was 250 Watt. The motor was set to move at a constant speed, which was 0.1 m/s in most cases, except when we tried the crawler uphill on a 20° slope, where it was set to 0.07 m/s. At this level of power and a speed of 0.1 m/s, we expected the crawler to develop a tractive force of up to 1.5 kN, for a maximum pull/weight ratio of 4.

The spikes of the first frame were only meant to pull the second frame forward and were comparatively small, designed for a maximum penetration depth of up to 15 cm. Their spike radius was 60 cm, and their spike diameter was 12 mm. The spikes of the second frame, intended to push the first frame, the blade, and the granular material forward, were designed for a maximum penetration depth of 50 cm, had a spike radius of 135 cm, and a spike diameter of 21 mm. Previous tests on beach sand had demonstrated that these spikes can easily sustain a draft of 2 kN and more.

The beach at the Bay of Palma had a well-defined slope along the waterline which increased from 12° at the waterline to 16° at the top edge, from where it continued inland almost flat for 25–30 m. The edge was 1.7 m above the waterline. The horizontal distance from the waterline to the edge was 6.8 m. The sand was moist, especially close to the waterline. No rain had been recorded in the week before the trials.

At the quarry, stone is broken, sorted by size, and stockpiled at various locations in the quarry. To allow access with heavy machinery, the larger stockpiles have an access ramp of 17°–20° on one side. All other sides form at the angle of repose of the material as it is dumped from the top. We tested on two different stockpiles: the stockpile with the finest material and another stockpile of coarse sand mixed with small stones, apparently what had been scraped off the ground after clearing another stockpile. The stockpile with the finest material had a 20° access ramp, which was uneven because of traffic by heavy machinery, and a smooth wall at an angle of repose of 40°. The pile of coarse sand had a slope angle of 30° and no ramp. It had an undulating surface because every dump of new material formed a new cone that partly overlapped with the old ones. The material at the quarry was moist but well-drained, 7 days or more after the last recorded rainfall, except for two trials: uphill on the 20° slope and contour following on the 40° slope were done one day after the last rain.

We tried the crawler in three different orientations along the slope of the different faces: following the contour, moving up diagonally to the line of steepest ascent, and moving uphill along the line of steepest ascent. When following the contour, vehicle roll matched the terrain slope while vehicle pitch was 0°. Moving uphill, vehicle roll was 0° while vehicle pitch matched the terrain slope. When moving diagonally uphill, both values were somewhere in between. We tried to let the crawler move 10 m or more during each trial. This was not always possible. For example, when moving uphill at the beach, the maximum possible path length was 7 m.

Table 2 offers a complete overview of all trials. Results were recorded on paper, in photos, and in videos. Vehicle angles were measured with the vehicle at rest. Electrical power was recorded at the power source. The trials were evaluated for whether the spikes penetrated the ground reliably, whether the crawler consistently moved forward, whether it moved straight (vehicle steering was not part of the trial, but a vehicle that won't move straight by itself tends to be more difficult to steer), whether the crawler successfully pushed sand, and whether it successfully cleared a path.

**RESULTS**

On the beach, the crawler performed mostly as intended. The spikes entered the sand reliably. The blade pushed sand at the intended peak force of 1.5 kN and cleared a path. When following the contour and when moving diagonally uphill, the vehicle veered slowly downhill, see Figure 2. When moving uphill, the path was straight.

At the quarry on the 20° ramp of fine material, the spikes entered the material reliably. When following the contour and when moving diagonally uphill, the blade pushed the material at the intended peak force of 1.5 kN and cleared a path, see the upper photo in Figure 3. The vehicle might have veered downhill, but the terrain was too uneven to say so for certain.

When moving uphill on the 20° ramp of fine material at the quarry, we decreased the motor speed to 0.7 m/s and increased the volume of material in front of the blade, increasing the draft to an estimated 2.5 kN, for a pull/weight ratio of 6. When the terrain was even, the crawler pushed the sand and cleared the path. When the terrain was uneven, one of the large spikes reached a thrust angle of 25° and lifted the vehicle. It could not continue. When we reduced the volume in front of the blade to an estimated draft of 2 kN and a pull/weight ratio to 5, the crawler resumed operations. Except for the described effect at an increased pull/weight ratio, the spikes penetrated the material reliably on a 20° slope, the material was pushed forward, and the path was cleared.

On the 30° stockpile of coarse material, the crawler did not move straight but slowly veered downhill. The large rear spikes penetrated reliably and pushed the blade with material forward. However, the small spikes in front repeatedly lifted the blade, such that the crawler could not continue, see Figure 4.

On the 40° wall of fine material, the crawler quickly veered downhill without pushing any material, see Figure 5.

Table 2. Trials with main observations. For the slope, the average inclination is given, which might not agree with the measured vehicle orientation. Terrain slope is given in degrees and percent. The main text uses degrees only.

| Location | Vehicle path, nr. of trials | Vehicle rotation | Observations |
|---|---|---|---|
| beach, flat | flat, 1 trial | pitch: 1° roll: 0° | Reliable spike penetration. The crawler moved straight, pushed all sand forward, and cleared a path. |
| beach, 12° = 21% slope | contour, 3 trials | pitch: 0° roll: 12° | Reliable spike penetration. The crawler pushed all sand forward and cleared a path. The vehicle had a slight tendency to veer downhill. |
| beach, 12°–16° = 21%–29% slope | diagonal, 5 trials | pitch: 8° roll: ~12° | |
| | uphill, 3 trials | pitch: 12°–16° roll: 0° | Reliable spike penetration. The crawler moved straight, pushed all sand forward, and cleared a path. |
| quarry, fine material, 20° = 36% slope | contour, 1 trial | pitch: 0° roll: 20° | Reliable spike penetration. The crawler moved mostly straight (difficult to assess in uneven terrain), pushed all sand forward, and cleared a path. |
| | diagonal, 2 trials | pitch: N/A roll: 7° | |
| | uphill, 2 trials | pitch: 20° roll: 0° | Reliable spike penetration. The crawler moved straight. On even terrain, at a draft of ~ 2.5 kN, the crawler pushed all sand forward and cleared a path. On uneven terrain, at a draft of ~ 2.5 kN, one large spike lifted the crawler, and it could not move forward. At a draft of ~ 2 kN, the crawler pushed sand forward and cleared a path. |
| quarry, coarse material, 30° = 58% slope | contour, 2 trials | pitch: 0° roll: 30° | Reliable penetration of the large spikes. The crawler did not move straight but slowly veered downhill. When the small spikes in front worked well, the crawler pushed all sand forward and cleared a path. Eventually, the small spikes in front lifted the blade such that the crawler could not continue. |
| | diagonal, 1 trial | pitch: 12° roll: 27° | |
| | uphill, 1 trial | pitch: 29° roll: 10° | Reliable penetration of the large spikes. The crawler moved straight, pushed all sand forward, and cleared a path. The small front spikes lifted the blade, but the crawler could continue. |
| quarry, fine material, 40° = 84% slope | contour, 1 trial | pitch: 5° roll: 40° | The crawler veered quickly downhill. |

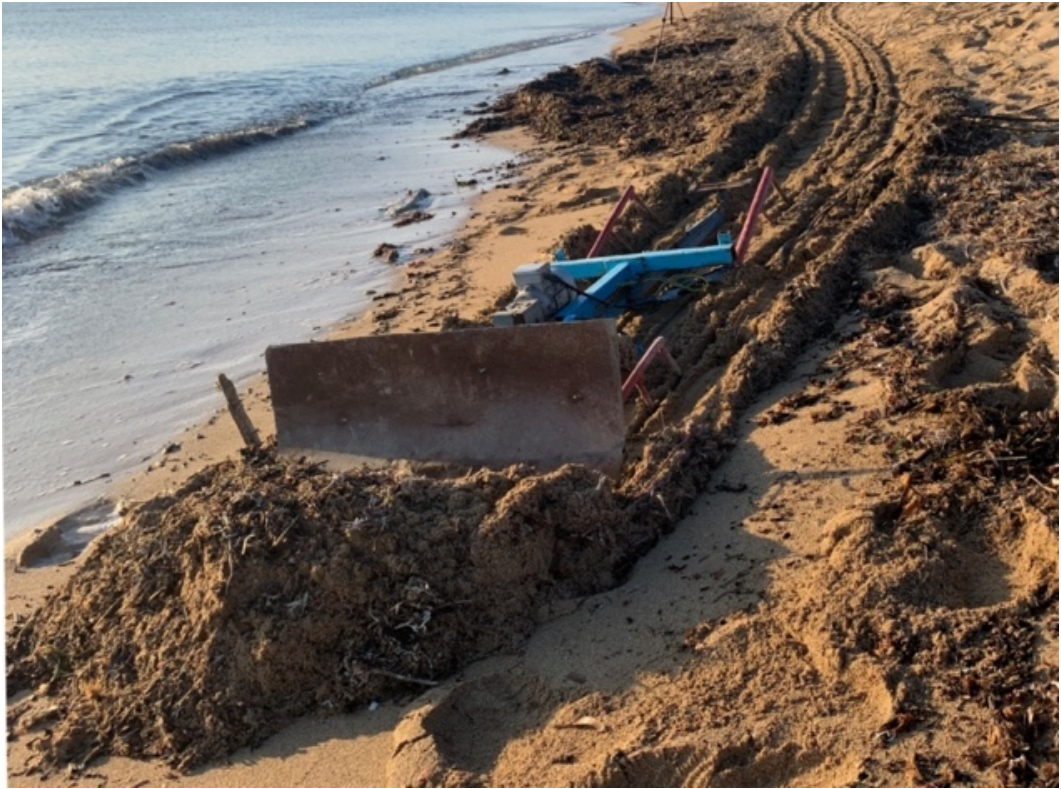

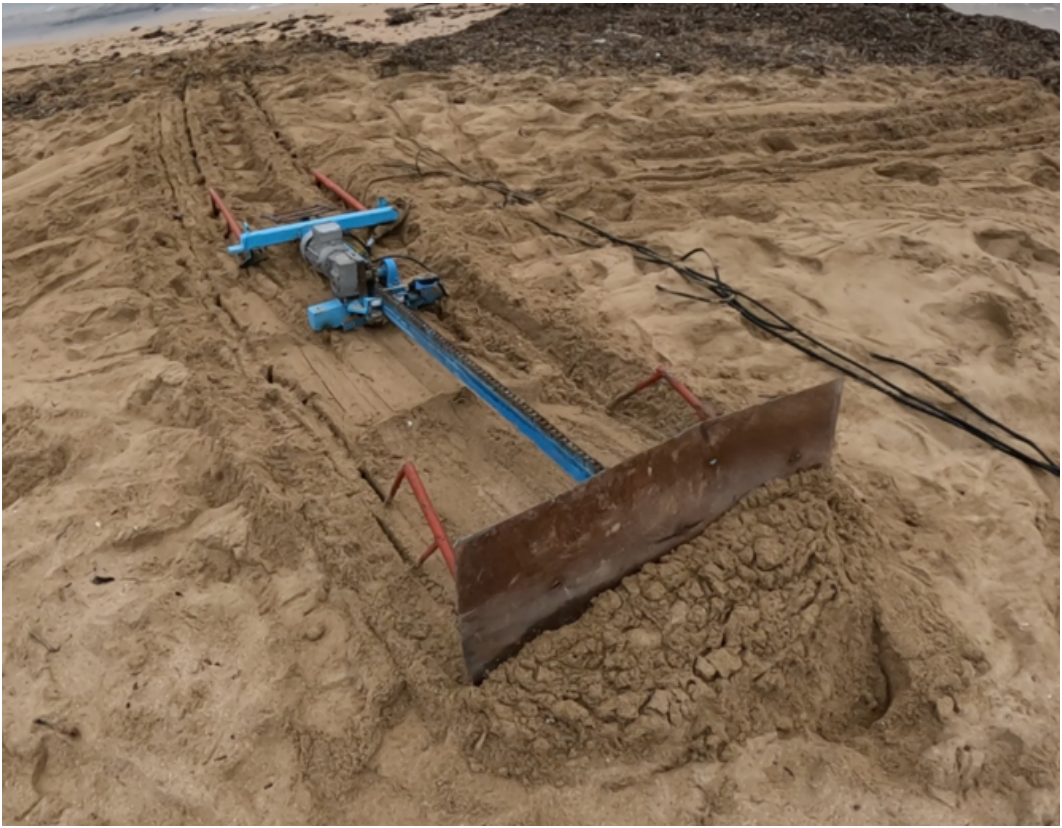

**Figure 2. Path clearing at the beach. Top: contour following. Bottom: diagonally uphill.**

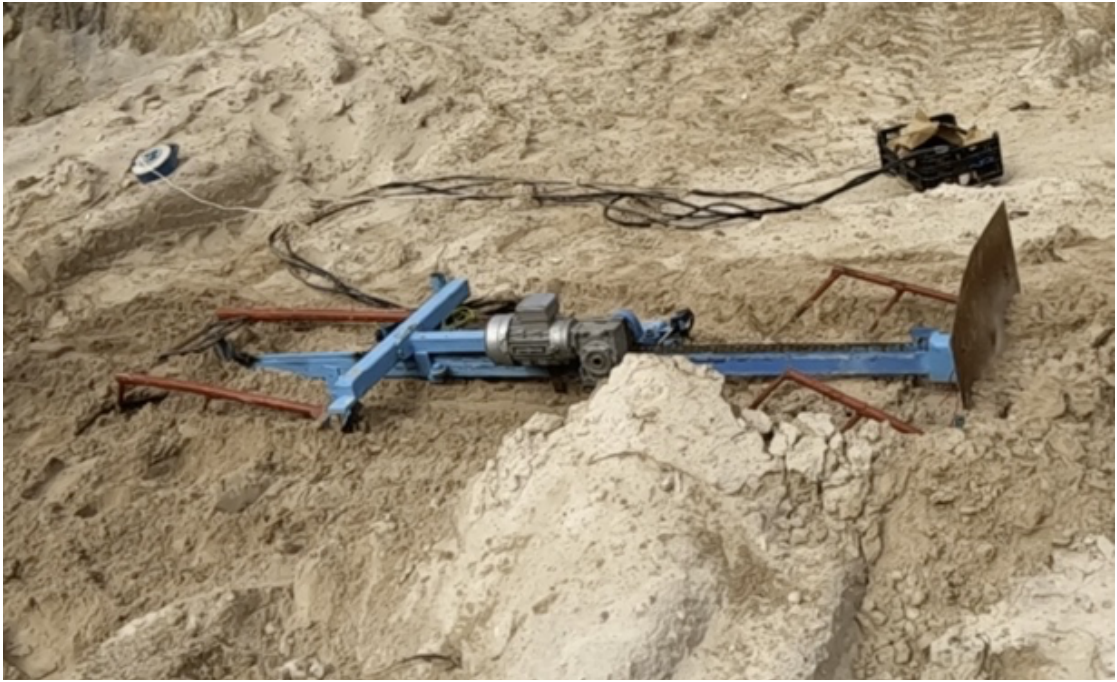

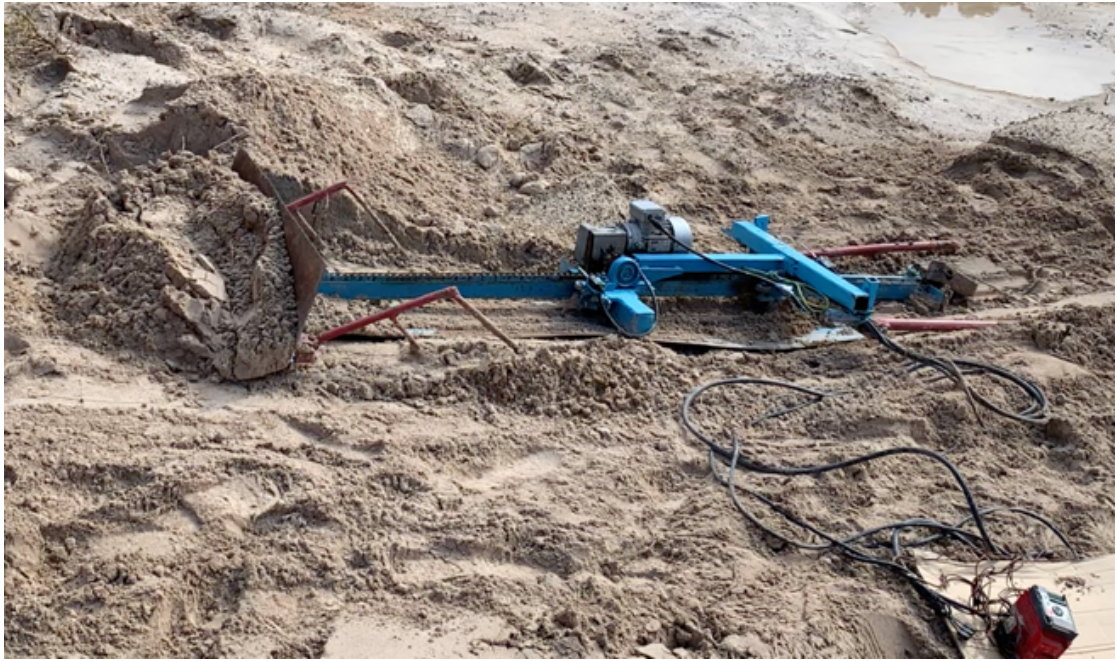

**Figure 3. Path clearing on the 20° ramp at the quarry. Top: contour following. The vehicle has successfully broken through a large accumulation of sand. Bottom: uphill against a draft of 2.5 kN.**

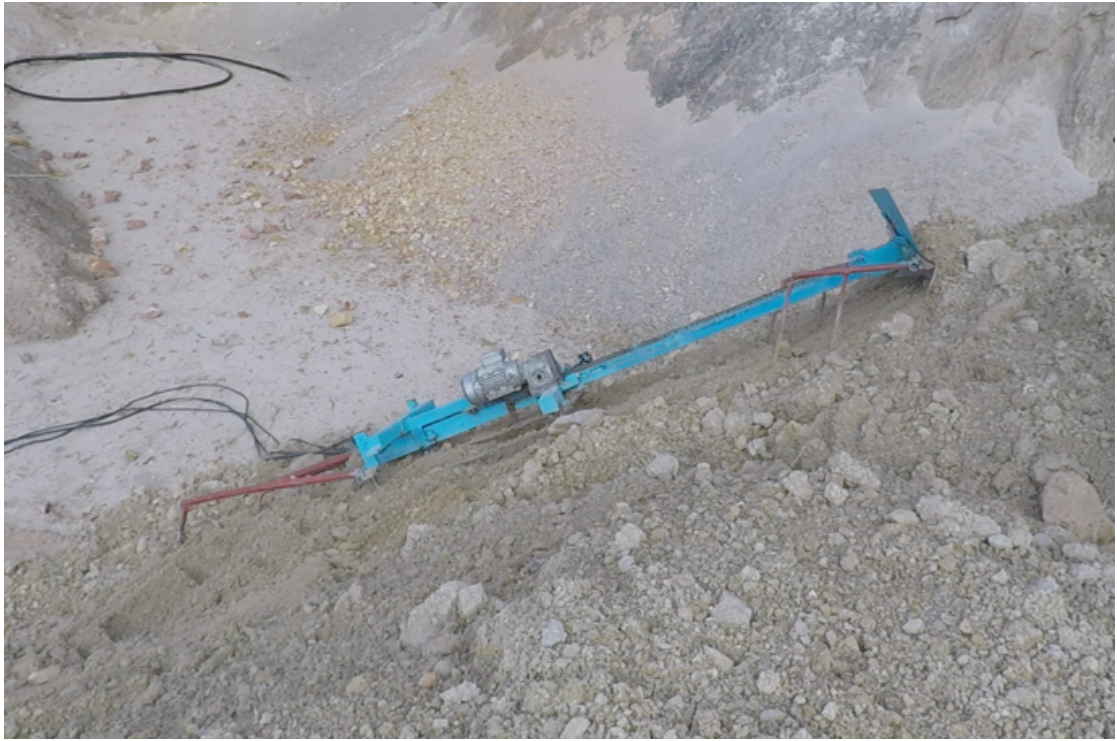
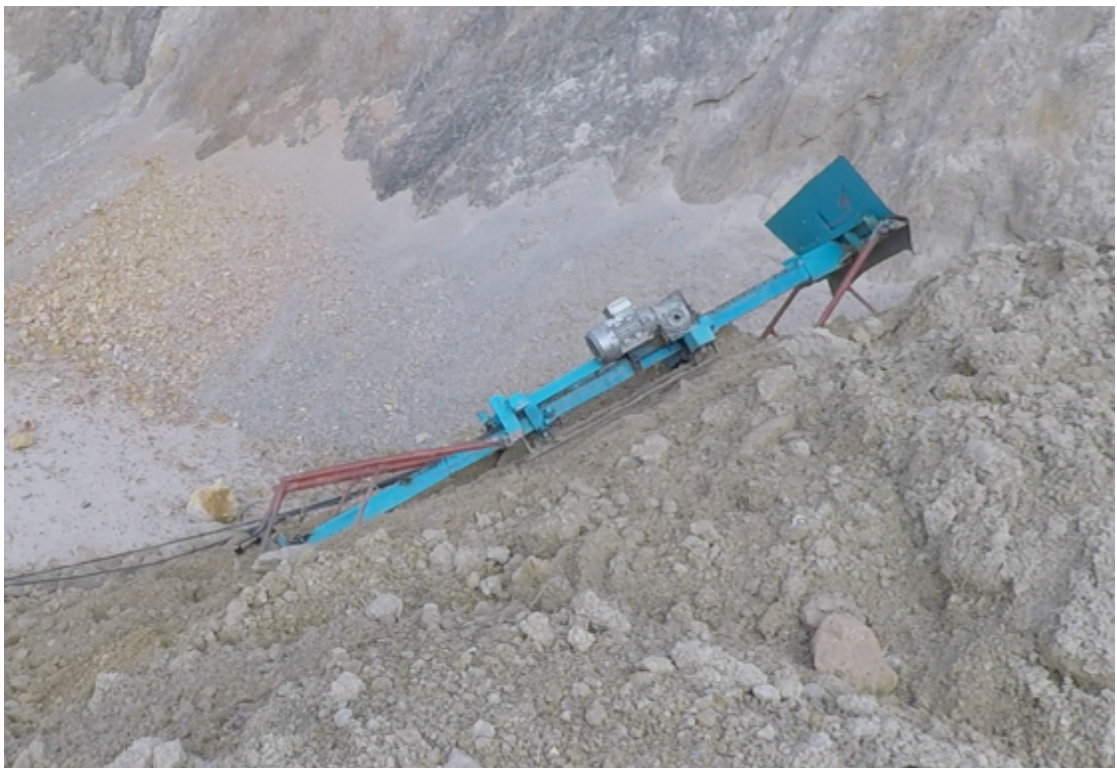

**Figure 4. Path clearing on a 30° slope at the quarry. Top: the crawler moves material. Bottom: the small spikes lift the blade; the vehicle cannot move forward.**

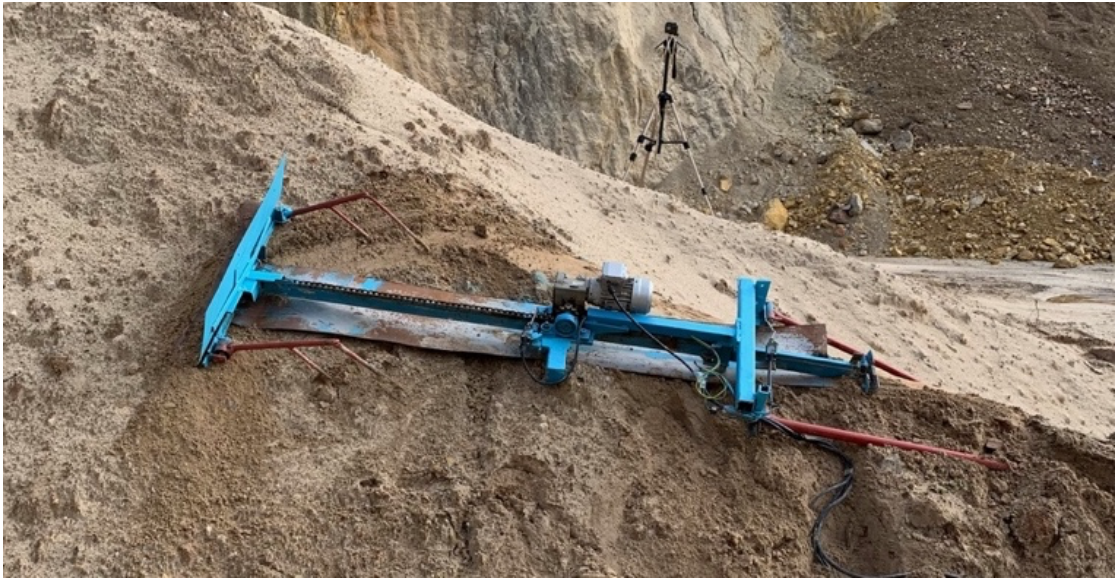
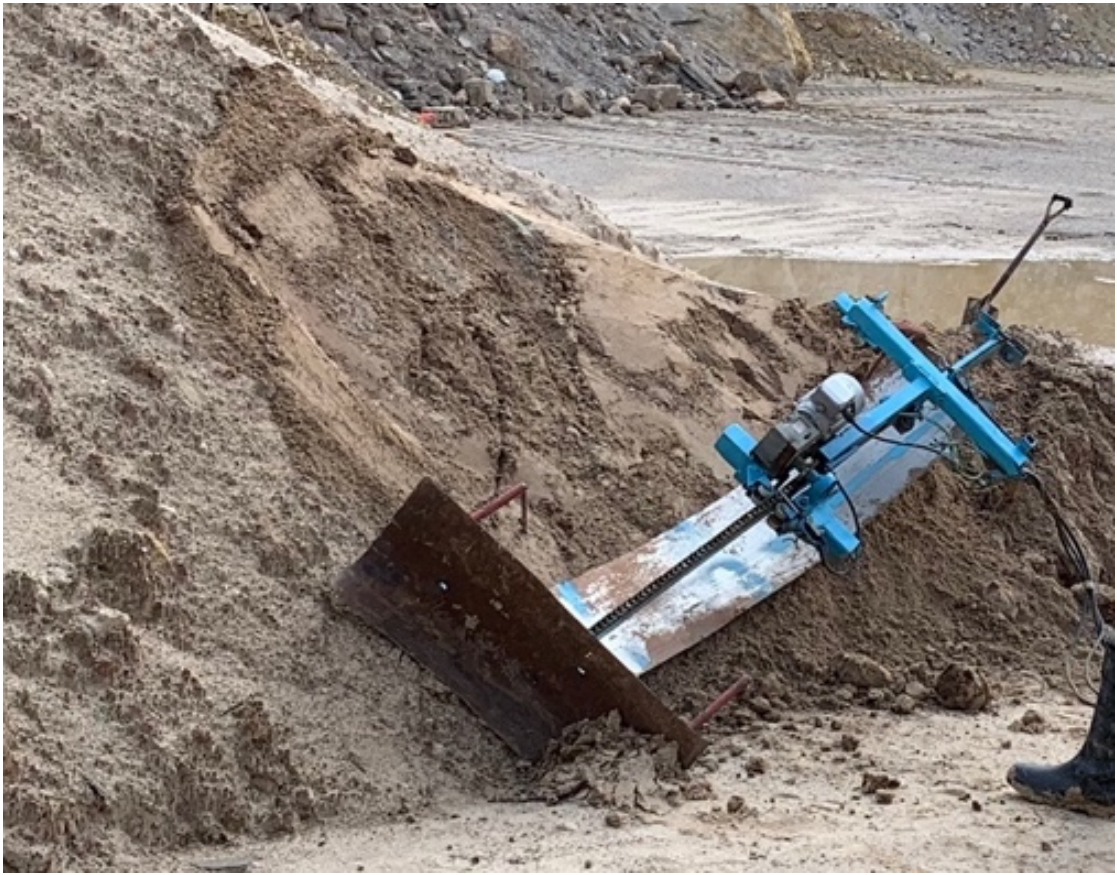

**Figure 5. On the 40° slope at the quarry, the crawler quickly veers downhill.**

## DISCUSSION AND CONCLUSION

We tested the performance of a crawler with spikes for traction on a side slope in two relevant environments: a beach and a quarry. While normal tires achieve an optimal tractive efficiency at a pull/weight ratio of 0.4 and cannot exceed a pull/weight ratio of 0.9, the tested crawler was designed to move earth at a pull/weight ratio of 4. A requirement for such a high pull/weight ratio is, that the thrust angle—the angle between the horizontal and a straight line between spike tip and hinge—does not exceed 15°. Otherwise, the pull on the spikes creates a lift at their hinges which exceeds vehicle weight.

During most trials, the thrust angle did not exceed 15°, and the vehicle safely reached the intended peak power and peak force. On the 20° slope we decreased the vehicle speed while keeping the power supply constant and increased the load at the blade, thereby increasing the pull/weight ratio to 6. The crawler performed well when the terrain was even. When the terrain was uneven, the thrust angle of one spike increased beyond 15° and lifted the machine, even though the other spike had the intended thrust angle of 15°.

We conclude that on granular materials the calculated pull/weight ratio can be exceeded, though not by much. We assume that the reason for this is, that in granular materials with little or no cohesion the center of force of the spike is not at the tip but further up, decreasing the effective thrust angle. We suggest that an actuator that prevents one spike to penetrate much deeper than other spikes could improve vehicle stability in uneven terrain.

We observe that the crawler performed generally well on slope angles of up to 20°. The spikes penetrated reliably, the crawler pushed material as intended and cleared a path. When following the contour or moving up diagonally, the vehicle veered slowly downhill, which requires adaptive path control.

At a 30° slope angle, the vehicle in its present configuration failed to operate as intended, chiefly because the front spikes were too small for the task (or the front of the vehicle was too light). The vehicle also tended to veer downhill. With some improvements to the design, the vehicle might be able to traverse a 30° slope angle. Whether it can also perform useful work while moving along the contour of or diagonally up a 30° slope needs to be further investigated.

While we have previously demonstrated that a similar vehicle can climb a 40° slope angle (Nannen et al., 2016), we find that in the present configuration, the vehicle cannot safely traverse a 40° side slope angle.

## ACKNOWLEDGEMENT

We thank the owners of Gravillera Son Chibetli, S.L. for their generous permission and support while working at their quarry.